\documentclass[letterpaper, 10 pt, conference]{ieeeconf}  
\usepackage{siunitx}
\usepackage{subcaption}
\IEEEoverridecommandlockouts
\usepackage[utf8]{inputenc}
\usepackage{graphicx}
\usepackage{hyperref}
\usepackage{amsmath,amssymb}
\usepackage{xspace}
\usepackage{tikz}

\usepackage{listings}
\usepackage{algorithm}
\usepackage{algpseudocode}

\usepackage{xcolor}

\title{\Huge {\bf Motion Planning Explorer}: Visualizing Local Minima using a Local-Minima Tree}
\date{May 2019}

\author{Andreas Orthey\thanks{The  authors  are  with  the  University of Stuttgart, Germany (e-mail: \{andreas.orthey, marc.toussaint\}@ipvs.uni-stuttgart.de)}, Benjamin Frész, Marc Toussaint}

\def\R{\mathbb{R}}
\def\Rpos{\R_{\geq 0}}
\def\Rstrictpos{\R_{>0}}
\def\N{\mathbb{N}}

\makeatletter
\newcommand{\algorithmicbreak}{\textbf{break}}
\newcommand{\Break}{\State \algorithmicbreak}
\makeatother

\def\simeq{\sim_{\fcost}}
\def\simeqproj{\sim_{\{\fcost,\pi_{k}\}}}

\def\path{\ensuremath{\mathbf{p}}}
\def\cost{c}
\def\fcost{f_{\cost}}

\def\q{\ensuremath{\mathbf{q}}}
\def\qprime{{\q}^{\prime}}

\def\qk{{\q}_{k}}

\def\pathprime{{\path}^{\prime}}

\def\tmax{t_{\text{max}}}
\def\deltaS{\delta_S}
\def\tree{T}

\def\free{\textnormal{f}}

\def\X{X}
\def\Xk{\X_{k}}
\def\Xkk{\X_{k+1}}

\def\siVelocity{\si{\meter\per\second}}

\def\Xf{\X_{\free}}

\def\x{x}

\def\planningspace{\left(\X,\phi\right)}
\def\planningproblem{\left(\Xf,\x_I,\x_G\right)}

\def\pathspace{P}

\def\xr{x_{\text{rand}}}
\def\xn{x_{\text{near}}}
\def\xw{x_{\text{new}}}
\def\G{\ensuremath{\mathbf{G}}}
\def\S{\ensuremath{\mathbf{S}}}
\def\Gk{\G_k}
\def\Gkk{\G_{k+1}}

\def\Skk{\S_{k+1}}

\makeatletter
\newcommand{\toprule}{\hrule height.8pt depth0pt \kern2pt} 
\newcommand{\midrule}{\kern2pt\hrule\kern2pt} 
\newcommand{\bottomrule}{\kern2pt\hrule\relax}
\newcommand{\algcaption}[2][]{%
  \refstepcounter{algorithm}%
  \toprule
  \textbf{{\raggedright\fname@algorithm~\thealgorithm}}\ #2\par 
  \midrule
}
\makeatother

\begin{document}
\maketitle
\thispagestyle{empty}
\pagestyle{empty}

\begin{abstract}


Motion planning problems often have many local minima. Those minima are important to visualize to let a user guide, prevent or predict motions. Towards this goal, we develop the motion planning explorer, an algorithm to let users interactively explore a tree of local-minima. Following ideas from Morse theory, we define local minima as paths invariant under minimization of a cost functional. The local-minima are grouped into a local-minima tree using lower-dimensional projections specified by a user. The user can then interactively explore the local-minima tree, thereby visualizing the problem structure and guide or prevent motions. We show the motion planning explorer to faithfully capture local minima in four realistic scenarios, both for holonomic and certain non-holonomic robots.

\end{abstract}


\section{Introduction}

In motion planning, we develop algorithms to move robots from an initial
configuration to a desired goal configuration. Such algorithms are essential for
manufacturing, autonomous flight, computer animation or protein folding
\cite{lavalle_2006}.

Most motion planning algorithms are black-box algorithms\footnote{We call an
algorithm a black-box algorithm whenever the internal mechanism is hidden from
the user \cite{black_box}.}.  A user inputs a goal configuration and the
algorithm returns a motion. In real-world scenarios, however, black-box
algorithms are problematic. Human users cannot interact with the algorithm. There is no way to guide or prevent motions. Humans users cannot visualize the internal mechanism of the algorithm. There is no intuitive way to understand or debug the algorithm. Human users cannot predict the outcome of the algorithm. There is no way for coworkers to avoid or plan around a robot. Black-box algorithms are therefore an obstacle for having robots move in a safe, predictable and controllable way. 

In an effort to make robotic algorithms visualizable, predictable and
interactive, we develop the motion planning explorer. Using the planning explorer, we
enumerate and visualize local minima. Using ideas from Morse theory \cite{morse_1934}, we define a local minimum as \emph{a path which is invariant under minimization of a cost functional}. To each local minimum we can associate an equivalence class, the equivalence class of all paths converging to the local minimum.

Using this equivalence relation, we utilize a fiber bundle construction --- a sequence of admissible lower-dimensional projections
\cite{orthey_2019} --- to organize the local-minima into a tree. Since the number
of leaves of this tree is usually countable infinite, we do not compute the tree explicitly, but
let users interactively explore the tree.

This local-minima tree is primarily a tool to visualize the problem
structure. However, we believe it to be more widely applicable. The tree is a
visual guide to the (topological) complexity of the problem \cite{smale_1987}.
The tree visualizes where a deformation algorithm 
\cite{zucker_2013} converges to. The tree
allows us to interact with the algorithm, useful for
factory workers guiding their robot or the control of computer avatars.  The tree can be used to give high-level instructions to a robot --- crucial when bandwidth is limited. The tree provides alternatives for efficient replanning \cite{brock_2002}. Finally, the tree can be a source of symbolic representations \cite{toussaint_2017}. 
\subsection{Contributions}

We make three original contributions

\begin{enumerate}

    \item We propose a new data structure, the local-minima tree, to enumerate and organize local minima
    
    \item We propose an algorithm, the motion planning explorer, which creates a local-minima tree from input by a user
    
    \item We demonstrate the performance of the motion planning explorer on realistic planning problems and on pathological environments
    
\end{enumerate}

Our algorithm requires, for each robot, the specification of a fiber bundle by a user. We can then handle any holonomic robotic system \cite{lavalle_2006} and provide a first generalization to non-holonomic systems. 

\begin{figure}
    \centering
    \includegraphics[width=0.44\linewidth]{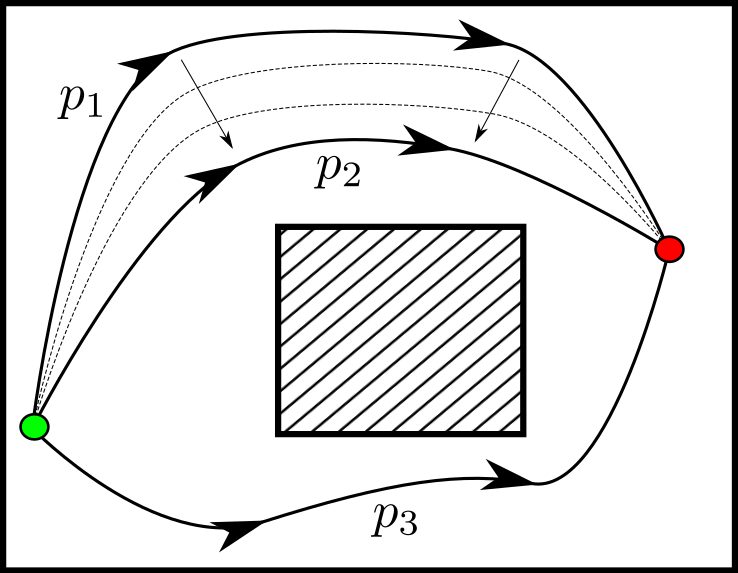}
    \includegraphics[width=0.52\linewidth]{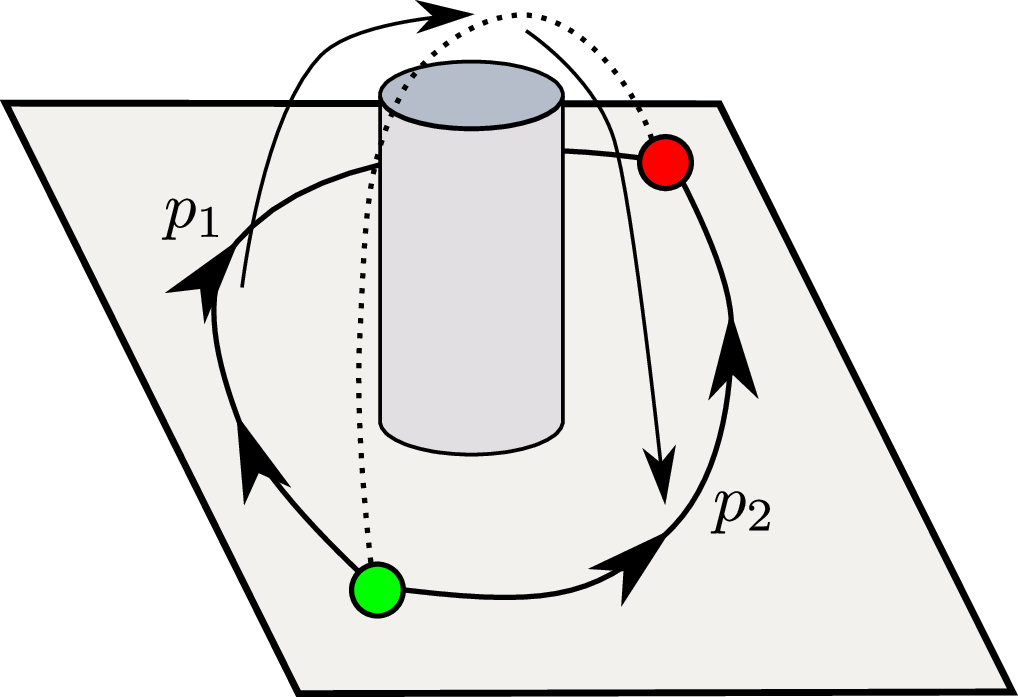}
    \caption{{\bf Left:} Homotopic paths in 2D. {\bf Right:} Distinct local minima which are homotopic in 3D.}
    \label{fig:homotopy}
\end{figure}

\section{Related Work}

We can visualize a motion planning problem by visualizing its decomposition. However, there is no clear consensus among researchers on the notion of decomposition. 

Often the problem is decomposed topologically \cite{farber_2008}. In a topological decomposition, we partition the pathspace into homotopy classes, sets of paths continuously deformable into each other \cite{munkres_2000} (See Fig. \ref{fig:homotopy} Left). We can compute homotopy classes by computing an H-signature of paths \cite{bhattacharya_2012} which counts, for each obstacle, the number of times a path crosses a line emanating from that obstacle. This can be generalized to higher dimensions, where we measure how often a path passes through holes in configuration space \cite{bhattacharya_2018}. 
If the configuration space is not too high-dimensional, we can also compute homotopy classes using simplicial complices \cite{pokorny_2016_ijrr} or lower-dimensional task projections \cite{pokorny_2016}.




Topological decompositions, however, do not adequately capture the intricate
geometry of configuration space constraints and are often computationally
inefficient. Many alternative definition have been proposed to obtain
computationally-efficient homotopy-like decompositions. Examples include digital
homotopy relations \cite{schmitzberger_2002}, $K$-order deformability
\cite{jaillet_2008} and convertibility of paths \cite{nieuwenhuisen_2004}. 

However, computationally-efficient homotopy decompositions fail to give a proper
pathspace partitioning. All previously named efficient decompositions are violating
the transitivity relation\footnote{Transitivity holds for a pathspace
decomposition if whenever a path $a$ is equivalent to a path $b$, and $b$ is
equivalent to a path $c$, then $a$ is equivalent to $c$.} and do not constitute
an equivalence relation. This makes it difficult to have clear lines of
demarcation between path subsets. It is also unclear how to visualize
overlapping path sets. 

We believe a more appropriate decomposition is the cost-function decomposition.
In a cost-function decomposition, we group paths together whenever \emph{they
converge under optimization to the same local minimum}. With such an approach, we can leverage optimization methods for computational efficiency and compute partitions of the pathspace. In Fig. \ref{fig:homotopy} (Right) we show a partition into two local-minima classes (ignoring minima wrapping around the obstacle).

The computation of local minima of cost functions belongs to the topic of
optimal motion planning \cite{karaman_2011}.  In optimal motion planning we like
to find the global cost-function minimum. Recently, several sampling-based
algorithms have been proposed which are asymptotically optimal, i.e. we will
find the global optimum if time goes to infinity \cite{karaman_2011}. Recent
extensions of those algorithms exploit graph sparsity \cite{dobson_2014},
improve upon convergence time \cite{janson_2015} and solve kinodynamic problems
\cite{li_2016}. 

However, most optimal planning algorithms will find only the global optimal
path, but not necessarily all local optimal paths. Our work differs by interactively computing
local minima and arranging them into a local-minima tree. To build the tree, we require fiber bundle simplifications of the configuration space \cite{orthey_2019}. Those simplifications help us to organize the local-minima of a planning problem and visualize its path space.

Visualization of path spaces is  closely related to topological data analysis and Morse theory. We briefly discuss those approaches and how they differ from our approach.

\subsubsection{Topological Data Analysis}

In topological data analysis \cite{carlsson_2009}, we use structures from algebraic topology (e.g. simplicial complices \cite{edelsbrunner_2010}) to model and visualize the topology of high-dimensional data. Our approach differs by computing paths directly (not modelling the topology) and by making the visualization interactive.

\subsubsection{Infinite-Dimensional Morse Theory}

The goal of infinite-dimensional Morse theory \cite{morse_1934} is studying solution spaces of optimization problems and investigating critical solution paths \cite{milnor_1963}. Our approach can be seen as applying Morse theory to motion planning while ignoring second-order behavior of the paths.

\section{Background}

Let $\planningspace$ be the \emph{planning space}, consisting of the configuration space $\X$ of a robot and the \emph{constraint function} $\phi:\X \rightarrow \{0,1\}$ which on input $x \in \X$ outputs zero when $x$ is constraint-free and one otherwise. We extend $\phi$ such that on input of subsets $U \subseteq \X$ outputs zero when at least one $x \in U$ is constraint-free and to one otherwise.
The constraint function defines the \emph{free configuration space} $\Xf = \{\x \in \X \mid \phi(\x) = 0\}$. 
Given an initial configuration $\x_I \in \Xf$ and a goal configuration $\x_G \in \Xf$, we are interested in finding a path in $\Xf$ connecting them. We call $\planningproblem$ a \emph{motion planning problem} \cite{orthey_2019}.


The space of solutions to a planning problem is given by its \emph{path space}. The path space $P$ is the set of continuous paths $\path: I \rightarrow \Xf$ from $I=[0,1]$ to $\Xf$ such that $\path(0)=x_I$ and $\path(1)=x_G$. We equip the pathspace $P$ with a cost functional $c: P \rightarrow \Rpos$ on $P$. Examples of cost functionals are minimum-length, minimum-energy, or maximum-clearance. 

\subsection{Admissible Fiber Bundles}

We can often simplify planning spaces using \emph{fiber bundles} \cite{orthey_2019}. A fiber bundle is a tuple $(\X, Y, \pi, \pi_\phi)$ consisting of a mapping
\begin{equation}
    \pi: \X \rightarrow Y
\end{equation}
which maps open sets to open sets and a mapping $\pi_\phi: \phi \rightarrow \phi_Y$, which 
map a planning space $\planningspace$ to a lower-dimensional space $\left(Y,\phi_Y\right)$. We
say that $\pi_\phi$ is \emph{admissible} if the admissibility condition $\phi_Y(y) \leq \phi(\pi^{-1}(y))$ holds for all $y \in Y$, whereby we call $\pi^{-1}(y)$ the \emph{fiber} of $y$ in $X$. We then call $\pi$ an admissible lower-dimensional projection, $\X$ the bundle space, and $Y$ the \emph{quotient space} of $X$ under $\pi$ \cite{lee_2010}.

Often it is advantageous to define chains of $K$ fiber bundles $(\X_k,\X_{k-1},\pi_k,\pi_{\phi_k})$ with admissible mappings
\begin{equation}
    \{\pi_k: \X_k \rightarrow \X_{k-1}\}_{k=1}^K
\end{equation}
such that $\X_K = \X$ and the constraint functions are admissible such that $\phi_{k-1}(x_{k-1}) \leq \phi_k(\pi_{k}^{-1}(x_{k-1}))$ for all $x_{k-1} \in \X_{k-1}$. Admissible fiber bundles have been shown to be a generalization of constraint relaxation, a source of admissible heuristic and can reduce planning time by up to one order of magnitude \cite{orthey_2019}.

There are multiple ways of simplifying a configuration space to construct a fiber bundle. We can often construct simpler robotic system by removing constraints, through nesting lower degree-of-freedom (dof) robots \cite{orthey_2018}, removing links \cite{bayazit_2005} or shrinking obstacles \cite{ferbach_1997}.

If the projection mappings $\pi$ and $\pi_\phi$ are obvious from the context, we will often denote the fiber bundle simply as $X \rightarrow Y$. As an example, we write $SE(2) \rightarrow \R^2$ for the car in Fig. \ref{fig:car}, whereby we mean that the car has been simplified by a nested disk. The nested disk is an abstraction of the car, removing the orientation. The mapping $\pi$ in that case maps position and orientation onto position, and $\phi_{\R^2}$ is zero whenever the disk is collision-free.

\section{Method\label{sec:method}}

In this section, we describe the \emph{local-minima tree}. First, we
define local minima as paths which are invariant under minimization of a cost functional. We then associate an equivalence class to each local minimum, consisting of all paths converging to the same local minimum. Using this equivalence relation, we then construct a
local-minima space. To visualize the local-minima space, we finally group local-minima
into a tree using the fiber bundle construction \cite{orthey_2019}. 

\subsection{Assumptions}

Let $\planningproblem$ be a motion planning problem, $\pathspace$
its path space, and $\cost: \pathspace \rightarrow \Rpos$ be a cost functional
on the pathspace. We assume that there exists a path optimization algorithm that we represent as a mapping $\fcost:
\pathspace \rightarrow \pathspace$, which takes any path and transforms the path into a path having a locally minimal cost. We make no further assumptions about the optimizer, such as that the output optimum is close to the initialization. Instead, our notion of path equivalence will be relative to a given $\fcost$. Further, we let  a user provide an admissible fiber bundle $\X_K \rightarrow \X_{K-1}
\rightarrow \cdots \rightarrow \X_0$ with $\X_K = \X$, which simplifies the configuration space $X$. The fiber bundle implicitly defines lower-dimensional projections $\pi_K,\cdots,\pi_1$.

\subsection{Local-Minima Space}

The minimization function $\fcost$ \emph{partitions}\footnote{A
partition of a set $X$ is a family of disjoint non-empty sets such that every
element of $X$ is in exactly one such set.} the pathspace. The partition is given by an equivalence relation we call path equivalence. Given the path-equivalence, we can construct the quotient of the pathspace
under path-equivalence, which we call the local-minima space.

Let us start by defining path-equivalence. If two paths converge, under the optimizer $\fcost$ of the cost $c$,
to the same path, we say they are path-equivalent. Formally, given two paths
$\path,\pathprime \in P$, we say that they are \emph{path-equivalent}, written as $\path \simeq \pathprime$, if 
\begin{equation} \fcost(\path) = \fcost(\pathprime) \end{equation}
It is straightforward to check that path-equivalence is an equivalence relation (i.e.
reflexive, symmetric, transitive). The optimizer $\fcost$ therefore
partitions the pathspace \cite{munkres_2000}. 

To better understand this partition, we construct the local-minima space
as the quotient space of all equivalence classes of $P$ under $\fcost$, denoted as
\begin{equation}
    Q = P/\simeq
\end{equation}
Elements of the space $Q$ are equivalence classes of paths. We will, however, \emph{represent} each equivalence class by the path which is invariant under minimization of the cost. We call those paths local minima.

To simplify matters, we will only consider \emph{simple} local minima. A simple local minimum is a local minimum without self-intersections. Simple paths are easier to compute and often capture all important local minima in a problem. However, we note that there are certain pathological cases, where non-simple paths are required to solve the problem \cite{orthey_2018}. 

\subsection{Sequential Projections of the Local-Minima Space}

To efficiently represent the local-minima space, we propose to sequentially partition the space using the fiber bundle projections.
This works as follows: Two distinct local minima of $Q$ are projected onto a quotient-space $X_k$ using the mapping $\pi_k$. We then consider them to be projection-equivalent, when, under minimization $\fcost$, they converge to the same path. 

More formally, given two local minima $\q, \qprime \in Q$, we
say that they are \emph{projection-equivalent}, written as $\q \simeqproj \qprime$, if 
\begin{align}
    \fcost(\pi_k(\q)) = \fcost(\pi_k(\qprime))
\end{align}
Projection-equivalence is again an equivalence relation and therefore partitions the
local-minima space. We denote the quotient of $Q$ under the
projection-equivalence as $Q_{K-1}$. We then iterate this process for each 
projection mapping. Thus, given an admissible fiber bundle $\X_K \rightarrow \X_{K-1}
\rightarrow \cdots \rightarrow \X_0$, we construct a sequence of local-minima
spaces $Q_K,\ldots,Q_0$ with $Q_K = Q$. In other words, the local-minima space $Q_{k-1}$ is obtained from $Q_{k}$
as the quotient-space 
\begin{equation}
  Q_{k-1} = Q_{k}/\simeqproj
\end{equation}
Elements of $Q_{k-1}$ are equivalence classes of local minima of $Q_k$. We will, however, \emph{represent} each equivalence class by the path to which all its elements (after projection) will converge to. 

\subsection{Local-Minima Tree}

Finally, we use the sequence of local-minima spaces to
construct the local-minima tree. The tree consist of all elements of $Q_0,\ldots,Q_K$ as nodes. Two nodes are connected by a
directed edge, if the first node is a local minimum $\q_k$ of $Q_k$, the
second node is a local minimum $\q_{k+1}$ of $Q_{k+1}$, and we have
$\fcost(\pi_{k+1}(\q_{k+1})) = \q_k$. Additionally, we add one empty-set root
node which is connected to every element of $Q_0$.

Note that a complete description of the local-minima tree is only possible in
trivial cases. In any real-world scenario, we can only hope to visualize
certain subsets of the tree.
\subsection{Examples}

To make the preceding discussion concrete, we visualize the local-minima tree for two examples. 

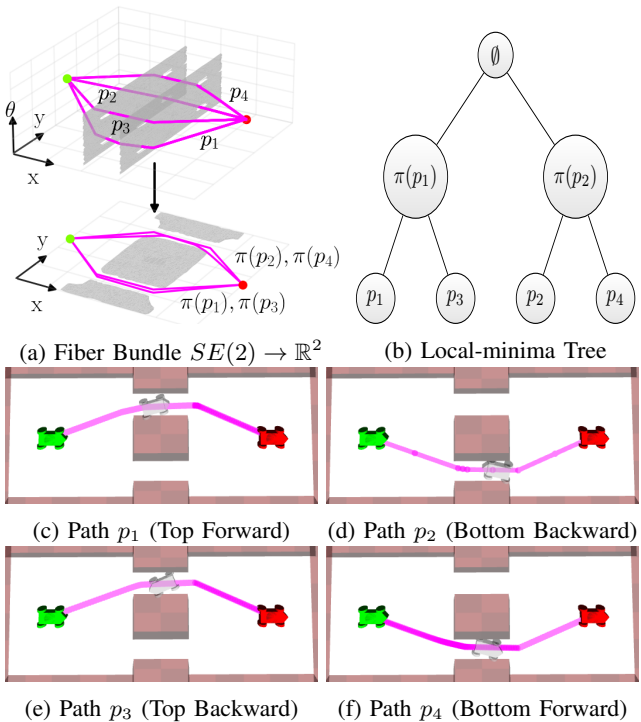
\begin{figure}[h!]
    \centering
     \begin{subfigure}[t]{0.52\linewidth}
         \centering
            \includegraphics[width=\linewidth]{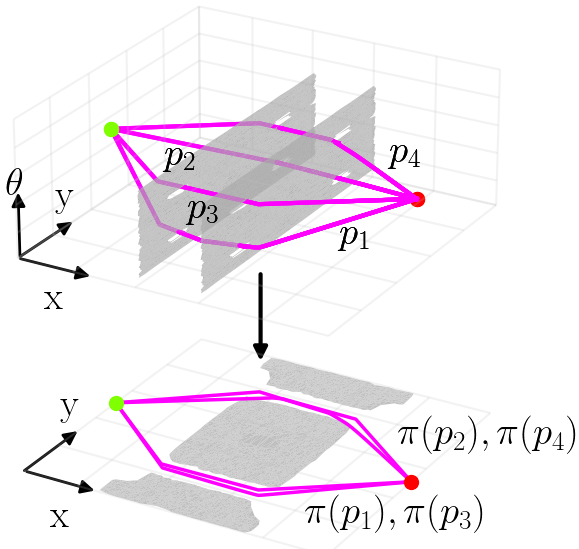}
         \caption{Fiber Bundle $SE(2) \rightarrow \R^2$}
         \label{fig:car_cspace}
     \end{subfigure}
     \begin{subfigure}[t]{0.45\linewidth}
     \centering
     \resizebox{\linewidth}{\linewidth}{
    \begin{tikzpicture}[
    level/.style={sibling distance=30mm/#1},
    level distance = 5em,
      every node/.style = {
        shape=circle, 
        rounded corners,
        draw, 
        align=center,
        top color=white, 
        bottom color=gray!20}]]
      \node {$\emptyset$}
        child { node {$\pi(p_1)$} 
            child { node {$p_1$}}
            child { node {$p_3$}}
        }
        child { node {$\pi(p_2)$}
            child { node {$p_2$}}
            child { node {$p_4$}}
        };
    \end{tikzpicture}
    }
    \caption{Local-minima Tree}
    \end{subfigure}

     \begin{subfigure}[b]{0.48\linewidth}
         \centering
            \includegraphics[width=\linewidth]{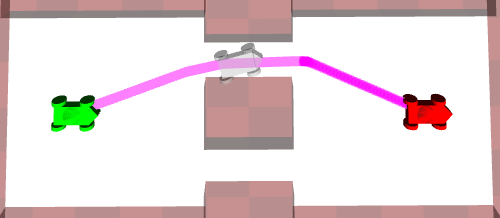}
         \caption{Path $p_1$ (Top Forward)}
         \label{fig:car_path1}
     \end{subfigure}
     \begin{subfigure}[b]{0.48\linewidth}
         \centering
            \includegraphics[width=\linewidth]{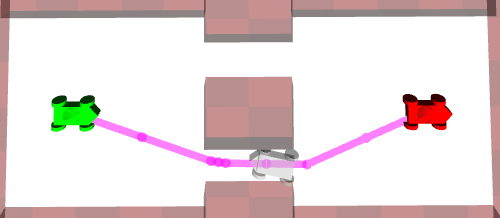}
         \caption{Path $p_2$ (Bottom Backward)}
         \label{fig:car_path2}
     \end{subfigure}     
     \begin{subfigure}[b]{0.48\linewidth}
         \centering
            \includegraphics[width=\linewidth]{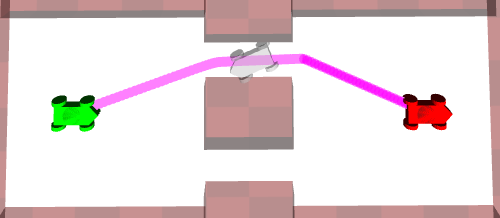}
         \caption{Path $p_3$ (Top Backward)}
         \label{fig:car_path3}
     \end{subfigure}     
     \begin{subfigure}[b]{0.48\linewidth}
         \centering
            \includegraphics[width=\linewidth]{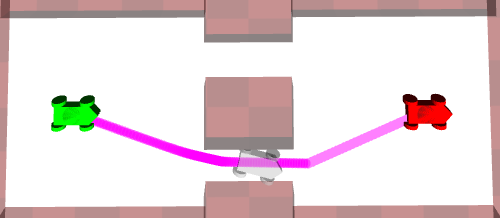}
         \caption{Path $p_4$ (Bottom Forward)}
         \label{fig:car_path4}
     \end{subfigure}

    \caption{Car in 2D with configuration space $SE(2)$. The planning problem can be decomposed into four parts.}
    \label{fig:car}
\end{figure}

First, we use a free-floating 3-dof planar car with fiber bundle $SE(2) \rightarrow \R^2$, which represents the removal of orientation by projection onto a circular disk. The environment is shown in Fig. \ref{fig:car} (c-f) and the fiber bundle is shown in Fig. \ref{fig:car}(a). The planning problem is to find a path to go from the green initial configuration to the red goal configuration. We observe that there are four simple local-minima, depending on if the car is going through the top or bottom slit, and going forward or backward. The two top slit paths are projection equivalent and we group them together. The same for the bottom paths. The local-minima tree is then shown in Fig. \ref{fig:car}(b). Note that we ignore non-simple local-minima which would occur when moving the car in a circle around the middle obstacle.

\begin{figure}[h!]
    \centering
 
    \begin{subfigure}[t]{0.48\linewidth}
         \centering
         \resizebox{\linewidth}{\linewidth}{             \includegraphics[width=0.9\linewidth]{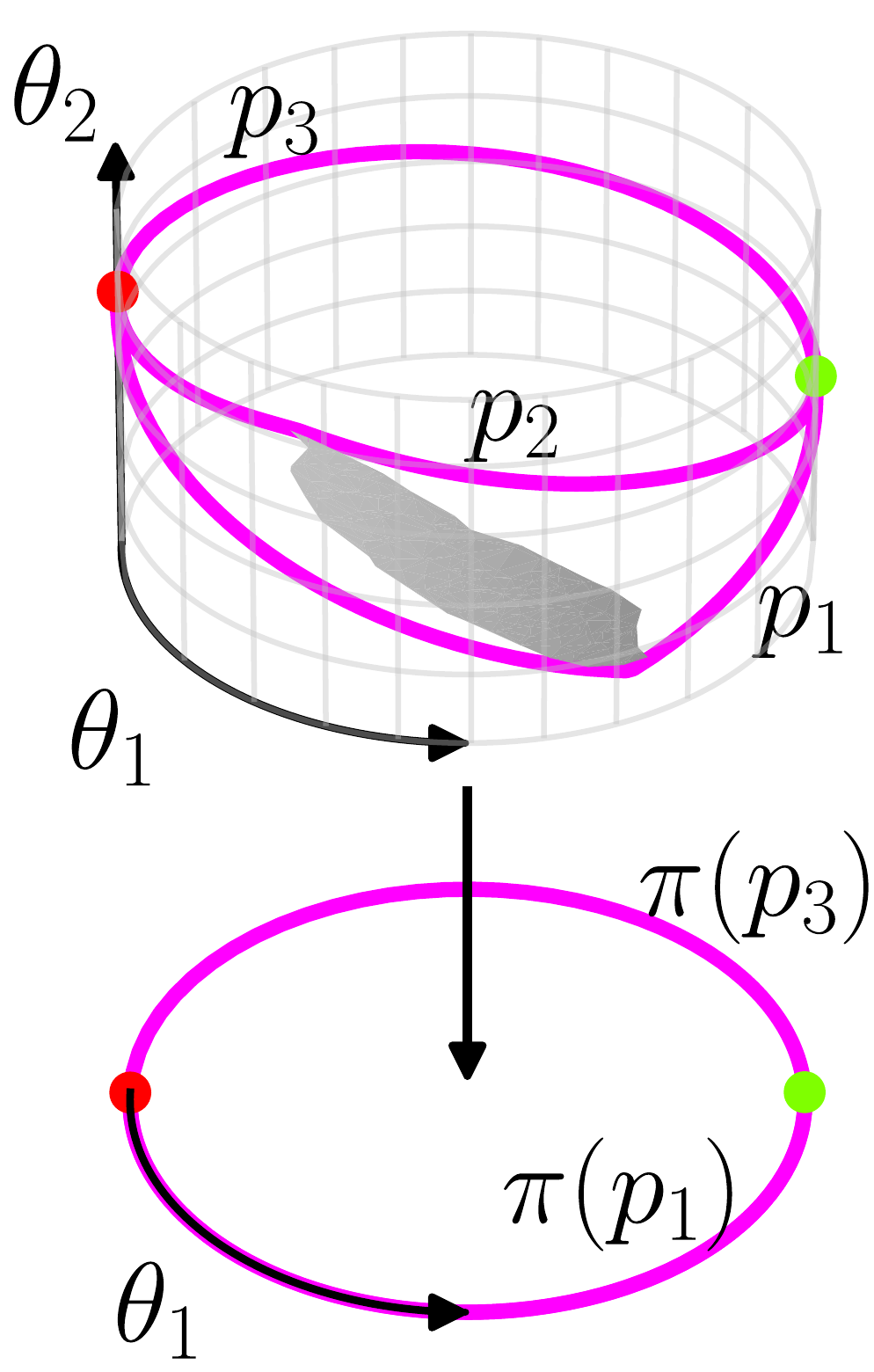}
         }
        \caption{Fiber Bundle $S^1\times \R^1 \rightarrow S^1$}
    \end{subfigure}  
    \begin{subfigure}[t]{0.48\linewidth}
     \centering
     \resizebox{\linewidth}{\linewidth}{
    \begin{tikzpicture}[      level/.style={sibling distance=30mm/#1},
      every node/.style = {
        shape=circle, 
        rounded corners,
        draw, 
        align=center,
        top color=white, 
        bottom color=gray!20}]
      \node {$\emptyset$}
        child { node {$\pi(p_1)$} 
            child { node {$p_1$}}
            child { node {$p_2$}}
        }
        child { node {$\pi(p_3)$}
            child { node {$p_3$}}
        };
    \end{tikzpicture}
    }
    \caption{Local-minima Tree}
    \end{subfigure}  
     \begin{subfigure}[b]{0.32\linewidth}
     \centering
        \includegraphics[width=\linewidth]{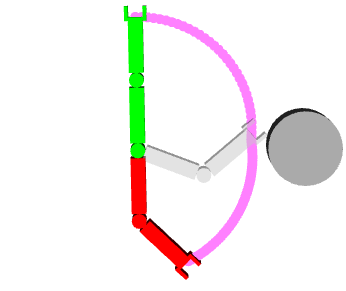}
     \caption{Path $p_1$ (Clockwise Below)}
     \label{fig:manip_p1}
    \end{subfigure}
     \begin{subfigure}[b]{0.32\linewidth}
     \centering
        \includegraphics[width=\linewidth]{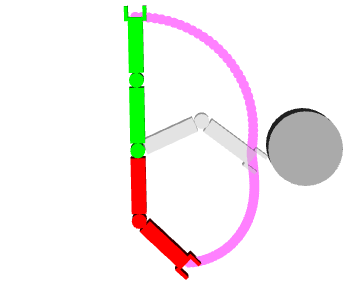}
     \caption{Path $p_2$ (Clockwise Above)}
     \label{fig:manip_p2}
    \end{subfigure}    
   \begin{subfigure}[b]{0.32\linewidth}
     \centering
        \includegraphics[width=\linewidth]{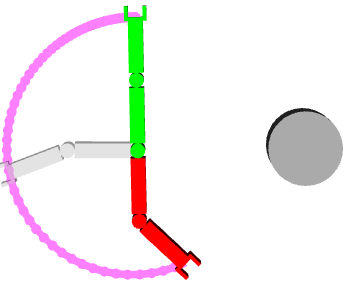}
     \caption{Path $p_3$ (Counter-Clockwise)}
     \label{fig:manip_p3}
    \end{subfigure} 
    \caption{2D Manipulator with environment which can be decomposed into three parts. Note that the three shown paths cannot be deformed into each other.}
    \label{fig:manip}
\end{figure}
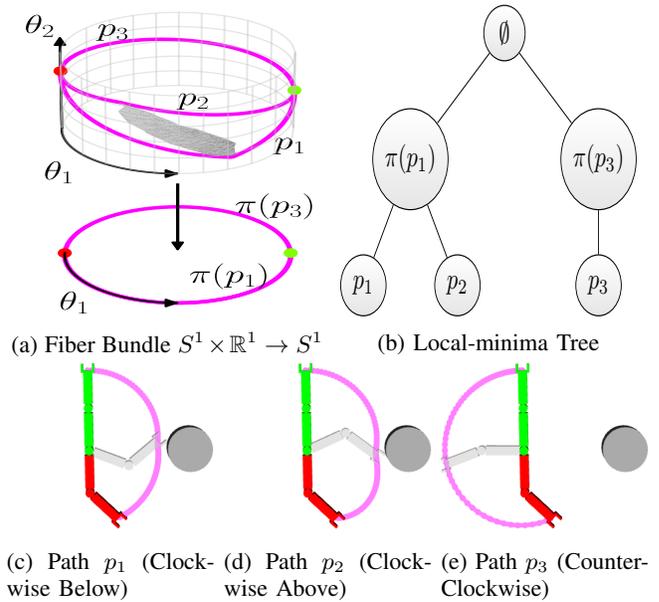

Second, we use a fixed-based 2-dof manipulator robot with fiber bundle $S^1 \times \R^1 \rightarrow S^1$ ($S^1$ is the circle space), which represents the removal of the last link. The environment is shown in Fig. \ref{fig:manip} (c-e) (obstacle in grey) with fiber bundle shown in  Fig. \ref{fig:manip}(a). There are three simple local-minima, two going clockwise below (c) and above (d) the obstacle, and one going counterclockwise (e). We group them according to their projection-equivalence as counterclockwise and clockwise, respectively. The local-minima tree is shown in Fig. \ref{fig:manip}(b). 

\lstset{language=C}
\alglanguage{pseudocode}
\def\Qn{Q_{new}}

\begin{figure}
    \centering
    \algcaption{MotionPlanningExplorer($x^I, x^G, \X_1,\ldots,\X_K, N, \tmax, \deltaS, \epsilon$)}
    \begin{algorithmic}[1]
      \State $T = \emptyset$
      \State $\G_1,\S_1,\ldots,\G_K,\S_K = \Call{InitRoadmaps}{\X_1,\ldots,\X_K}$
      \While{True}
      \State $\qk = \Call{SelectLocalMinima}{T}$\label{alg:explorer:selectlocalminima}
        \State $\Call{UpdateMinimaTree}{T, \qk, \Gk, \Gkk, \Skk}$\label{alg:explorer:updateminimatree}
      \EndWhile
     \State \Return $\emptyset$
    \end{algorithmic}
    \label{alg:explorer}
  \bottomrule
  \bigskip
  \algcaption{UpdateMinimaTree($\tree, \qk, \Gk, \Gkk, \Skk$)}

    \begin{algorithmic}[1]
        \While{$\neg\Call{ptc}{\tmax}$}
          \State $\Call{GrowRoadmap}{\Xkk, \qk, \Gk, \Gkk, \Skk}$\label{alg:updateminimatree:growroadmap}
        \EndWhile
        \State $\{q_1,\cdots,q_M\} \leftarrow \Call{EnumeratePaths}{\Skk}$\label{alg:updateminimatree:enumeratepaths}
        \State $\Qn \leftarrow \emptyset$
        \For{each $\q$ in $\{q_1,\cdots,q_M\}$ }\label{alg:updateminimatree:forstart}
             \If{$\neg$\Call{MinimaExists}{$\q,\Qn$}}\label{alg:updateminimatree:minimaexists}
                \State $\Qn \leftarrow \Qn \cup \q$
            \EndIf
            \If{$\Call{size}{\Qn} \geq N$}
                \Break
            \EndIf
        \EndFor\label{alg:updateminimatree:forend}
    \State \Call{AddMinimaToTree}{$\Qn, \tree$}\label{alg:updateminimatree:addminimatotree}
    \end{algorithmic}\label{alg:updateminimatree}
\bottomrule
  \bigskip
  \algcaption{MinimaExists($\q, \Qn$)}
  \begin{algorithmic}[1]

    \For{each $\qprime$ in $\Qn$ }
        \If{\Call{IsVisible}{$(\q,\qprime)$}}
           \State \Return True
        \EndIf
    \EndFor
   \State \Return False
  \end{algorithmic}
  \label{alg:instack}
  \bottomrule
  \bigskip
  \algcaption{GrowRoadmap($\Xkk, \qk, \Gk, \Gkk$)}
  \begin{algorithmic}[1]
    \State $\xr \gets \Call{SampleFiber}{\Xkk, \Gk, \qk, \epsilon}$\label{alg:growroadmap:samplefiber}
    \State $\xn \gets \Call{Nearest}{\xr,\Gkk}$\label{alg:growroadmap:nearest}
     \State $\xw \gets \Call{Connect}{\xn,\xr,\Gkk}$\label{alg:growroadmap:connect}
     \State $\Skk \gets \Call{AddConditional}{\xn, \Skk, \deltaS}$\label{alg:growroadmap:addconditional}
  \end{algorithmic}
  \label{alg:growroadmap}
  \bottomrule
\end{figure}

\section{Algorithm\label{sec:algorithm}}

To compute the local-minima tree, we develop the \emph{motion
planning explorer} (Algorithm  \ref{alg:explorer}). The motion planning explorer takes as input a planning
problem $\planningproblem$, a minimization method $\fcost$
and a fiber bundle represented as a sequence of quotient spaces $X_0,\cdots,X_K$ with $X_K = X$. The explorer depends on four
parameters, namely $N \in \N$, the maximum number of local-minima to display,
$\tmax \in \Rstrictpos$, the maximum time to sample in one iteration, $\deltaS \in \Rstrictpos$, the fraction of space to be visible for the underlying sparse roadmap and $\epsilon \in \Rpos$, the $\epsilon$-neighborhood of a local minimum to sample. Given the input, we return a browsable local-minima tree $T$. A user can navigate this tree by clicking
on local minima and by collapsing or expanding the minimum, similar to how one
navigates a unix directory structure. 

Our algorithm consists of an alternation of two phases. In phase one (Line \algref{alg:explorer}{alg:explorer:selectlocalminima}), a human
user can navigate the local-minima tree and select one local minima. In the
beginning the user has only one choice, selecting the root node (the empty-set
minimum). In the second phase (Line \algref{alg:explorer}{alg:explorer:updateminimatree}), the user presses a button and the algorithm
uses the selected local minimum $\q_k$ on $Q_k$ to find all local minima on
$Q_{k+1}$ which, when projected, would be equivalent to $\q_k$. For each local
minimum we find, we add a directed edge from $\q_k$ to the local minimum. Note that we construct the local-minima tree in a top-down fashion, which differs from the bottom-up description in Sec. \ref{sec:method}. This construction is more computationally efficient, but we might create \emph{spurious local-minima}, which are local minima which do not have any children. In other words, there are no local-minima which, when projected, would be equivalent to the spurious local minimum. This
second phase is run for a predetermined maximum timelimit $t_{max}$, and can be
run multiple times until the user has found sufficiently many local minima.

\begin{figure}[ht!]
    \centering
{%
\setlength{\fboxsep}{0pt}%
\setlength{\fboxrule}{1pt}%
\fbox{\includegraphics[width=\linewidth]{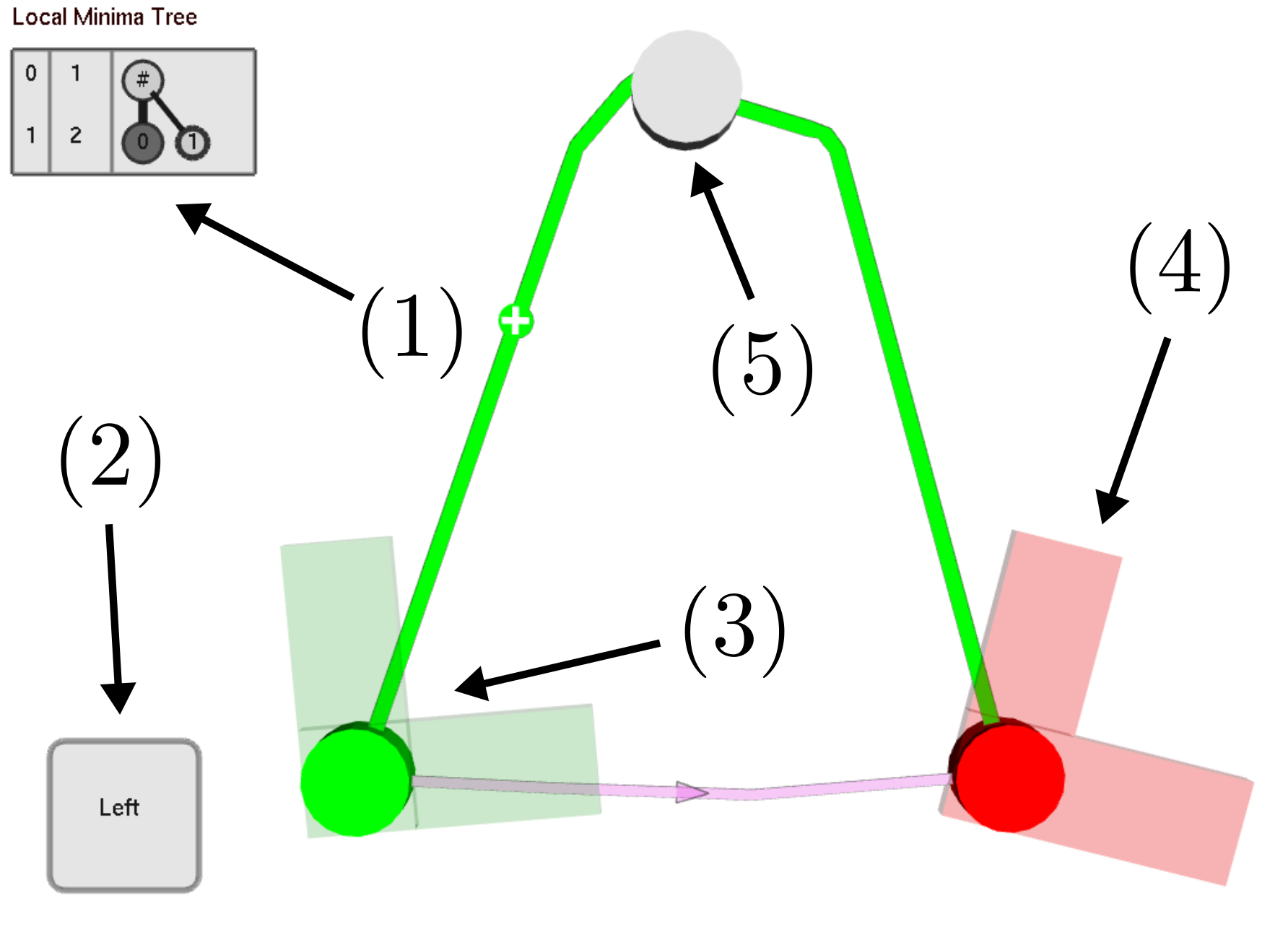}}%
}%
    \caption{{\bf Motion Planning Explorer GUI}: (1) local-minima tree depiction. Columns show fiber bundle level, number of nodes on level and nodes of tree, respectively. (2) last button pressed by user. (3) initial (green) configuration on quotient-space (non-transparent disk) and on bundle space (transparent). (4) same for goal (red) configuration. (5) local minima selected by user and highlighted in tree.}
    \label{fig:explorer}
\end{figure}

For phase one of the explorer, we develop a graphical user interface (GUI). The
GUI is shown in Fig. \ref{fig:explorer}, where we show the local minima
tree (1), the last button pressed (2), the initial configuration (3), the goal
configuration (4), and a configuration along a local minimum (environment is
hidden to remove distractions). Pressing the button
\texttt{left} or \texttt{right} switches local minima on the same level.
Pressing \texttt{up} collapses the current local minimum and displays the local minimum on the next lower-dimensional quotient space, which is obtained by projection and subsequent optimization of the current local minimum. Pressing \texttt{down} expands the current local minimum. Pressing
the button \texttt{u} executes the current local minimum path by sending it to the robot
and pressing the button \texttt{w} starts the search for more local minima.

In the second phase (Algorithm \ref{alg:updateminimatree}) we update the minima tree by performing two steps. First, we take the selected local minimum
$\q_k$ on $Q_k$ and grow a sparse graph $\Skk$ on the space $\Xkk$ (or $\X_K$ if
$k=K$) biased towards $\q_k$. Second, we  compute up to $N$ local-minima from the sparse graph $\Skk$. We first describe both steps in the case of a holonomic robot, and then describe the modifications in the non-holonomic case.

In the first step, we grow the sparse graph $\Skk$ on $\Xkk$ for up to $\tmax$ seconds (or some other Planner Terminate Condition (PTC)). The algorithm $\texttt{GrowRoadmap}$ (Line \algref{alg:updateminimatree}{alg:updateminimatree:growroadmap}) is further detailed in Algorithm \ref{alg:growroadmap}, which closely follows the Quotient-Space roadMap Planner
(QMP) algorithm \cite{orthey_2018}. It differs from QMP by computing both a dense graph $\Gkk$ and a sparse graph $\Skk$. To build the graphs, we first sample a configuration on the graph $\Gk$ biased towards an $\epsilon$-neighborhood of $\qk$. This configuration indexes a fiber through the inverse mapping $\pi^{-1}_k$. We then sample this fiber to obtain a configuraton on $\Xkk$ (Line \algref{alg:growroadmap}{alg:growroadmap:samplefiber}), compute the nearest configuration on $\Gkk$ (Line \algref{alg:growroadmap}{alg:growroadmap:nearest}) and connect if possible (Line \algref{alg:growroadmap}{alg:growroadmap:connect}). The new configuration is then conditionally added to the sparse graph (Line \algref{alg:growroadmap}{alg:growroadmap:addconditional}). Our implementation utilizes previous work from the Sparse Roadmap Spanners
(SPARS) algorithm\footnote{In particular, we add a configuration to the sparse roadmap
whenever the configuration increases visibility, increases connectivity or
constitutes a useful cycle.} \cite{dobson_2014}. The sparse graph $\Skk$ utilizes the parameters $\deltaS$ which determines the maximum visibility radius of a configuration. This method biases sampling towards paths which, when projected onto $\Xk$, will be projection-equivalent to $\q_k$. If we find a path not projection-equivalent to $\q_k$, we ignore the path.

In the second step, we enumerate $M \leq 2N$ paths on
$\Skk$ (Line \algref{alg:updateminimatree}{alg:updateminimatree:enumeratepaths}). Those paths are found using a depth-first graph search on $\Skk$. For each path
found, we use the optimizer $\fcost$ to let the path converge to the
nearest local minimum. We then try to add this path to a set of local minima paths (Line \algref{alg:updateminimatree}{alg:updateminimatree:forstart} to  \algref{alg:updateminimatree}{alg:updateminimatree:forend}). The path is added if it is
not visible from any path in the set (Line \algref{alg:updateminimatree}{alg:updateminimatree:minimaexists}). We implement the visibility function following the algorithm by \cite{jaillet_2008}. Another option would be to compute a distance between two paths. However, we found this to not work well with the particular minimization method we used in the demonstrations. Note that other minimization methods might require different methods to check convergence.

In the case of a non-holonomic robot, we replace the function $\texttt{GrowRoadmap}$ using an iteration of kinodynamic RRT \cite{kuffner_2000}. We then populate the sparse graph only with the current shortest path to the goal. This allows us to find a dynamically feasible path given a geometrically feasible path on the quotient space (using the path bias through $\qk$). However, for this work we did not implement an optimizer for dynamical systems and therefore can only return a single non-optimal path. In future work we need to use a sparse optimal graph spanner for kinodynamic systems like \cite{li_2016} and dynamical optimization functions for non-holonomic systems like \cite{lamiraux_2004}.

Once all simple paths have been enumerated, and the local minima saved, we stop the phase, add all found local minima to the
local-minima tree (Line \algref{alg:updateminimatree}{alg:updateminimatree:addminimatotree}
) and display them to the user in the GUI. Then we return to phase one.

The motion planning explorer has been implemented in C++ and uses the Klampt
library \cite{hauser_2016} for simulation and visualisation, and the Open Motion
Planning Library (OMPL) \cite{sucan_2012} for roadmap computation and fiber
bundle projection.  The implementation is freely available at
\href{https://github.com/aorthey/MotionPlanningExplorerGUI}{github.com/aorthey/MotionPlanningExplorerGUI}.

\begin{figure*}[ht!]
    \centering
    \includegraphics[width=0.32\textwidth]{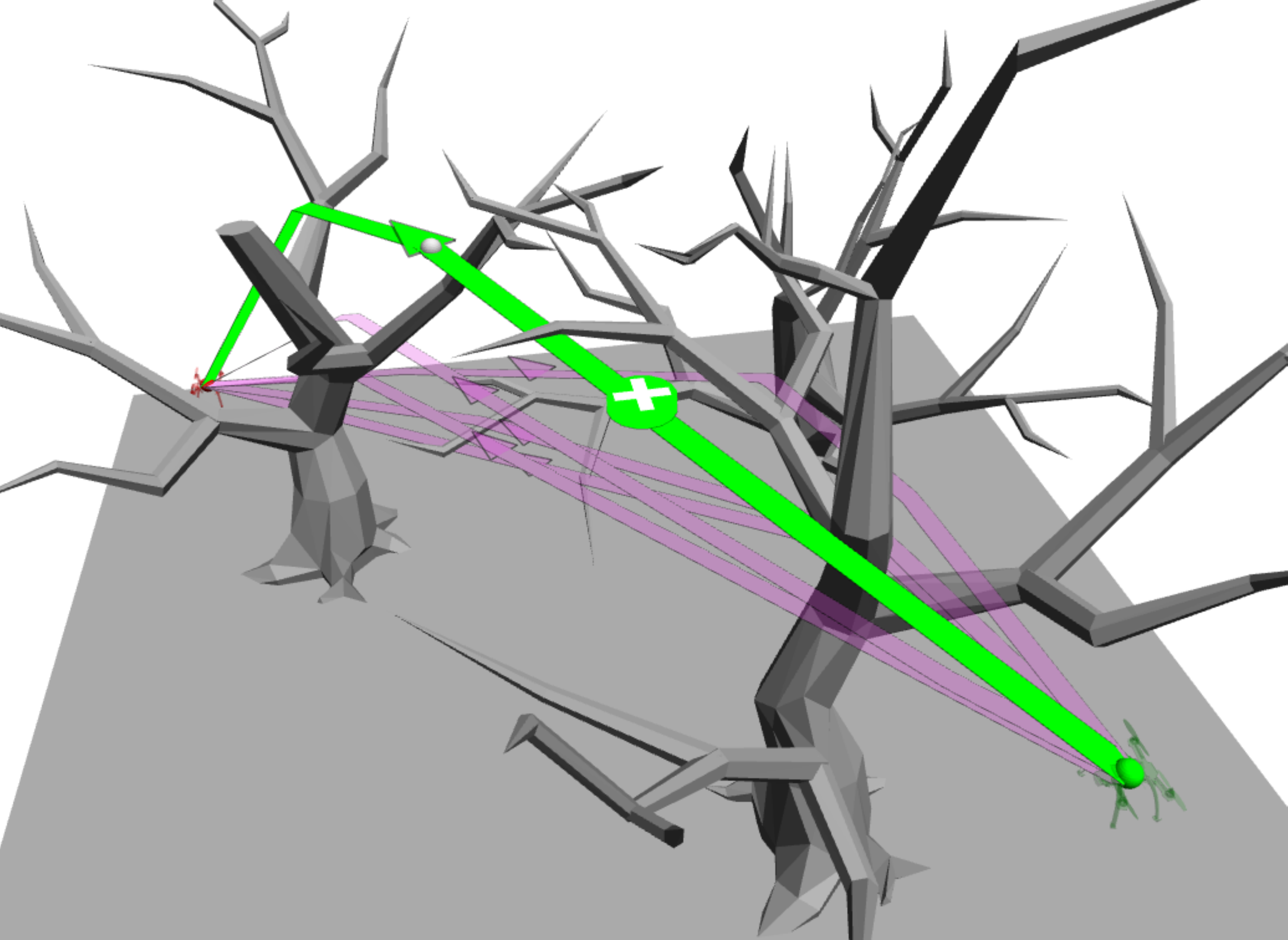}
    \includegraphics[width=0.32\textwidth]{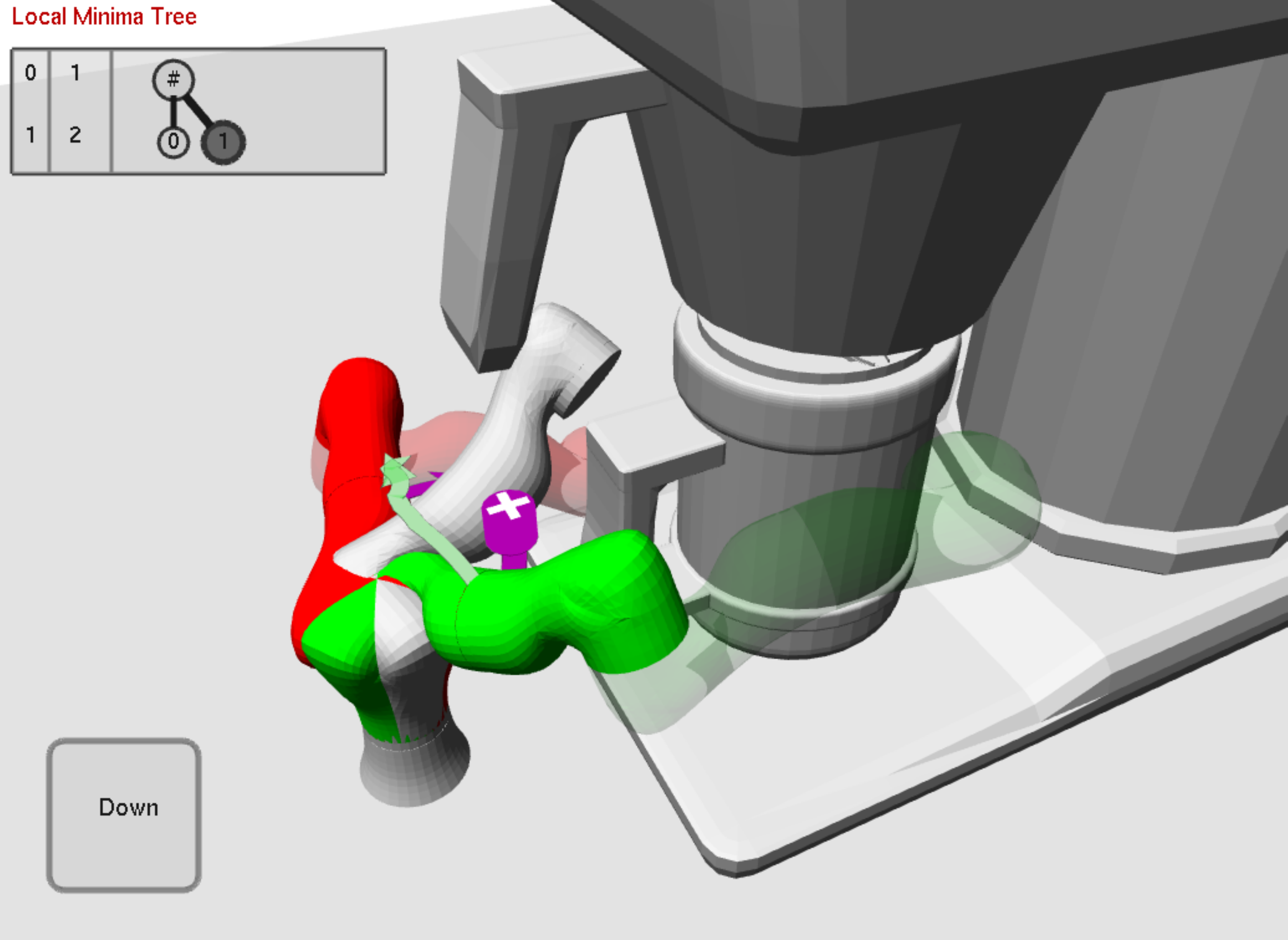}
    \includegraphics[width=0.32\textwidth]{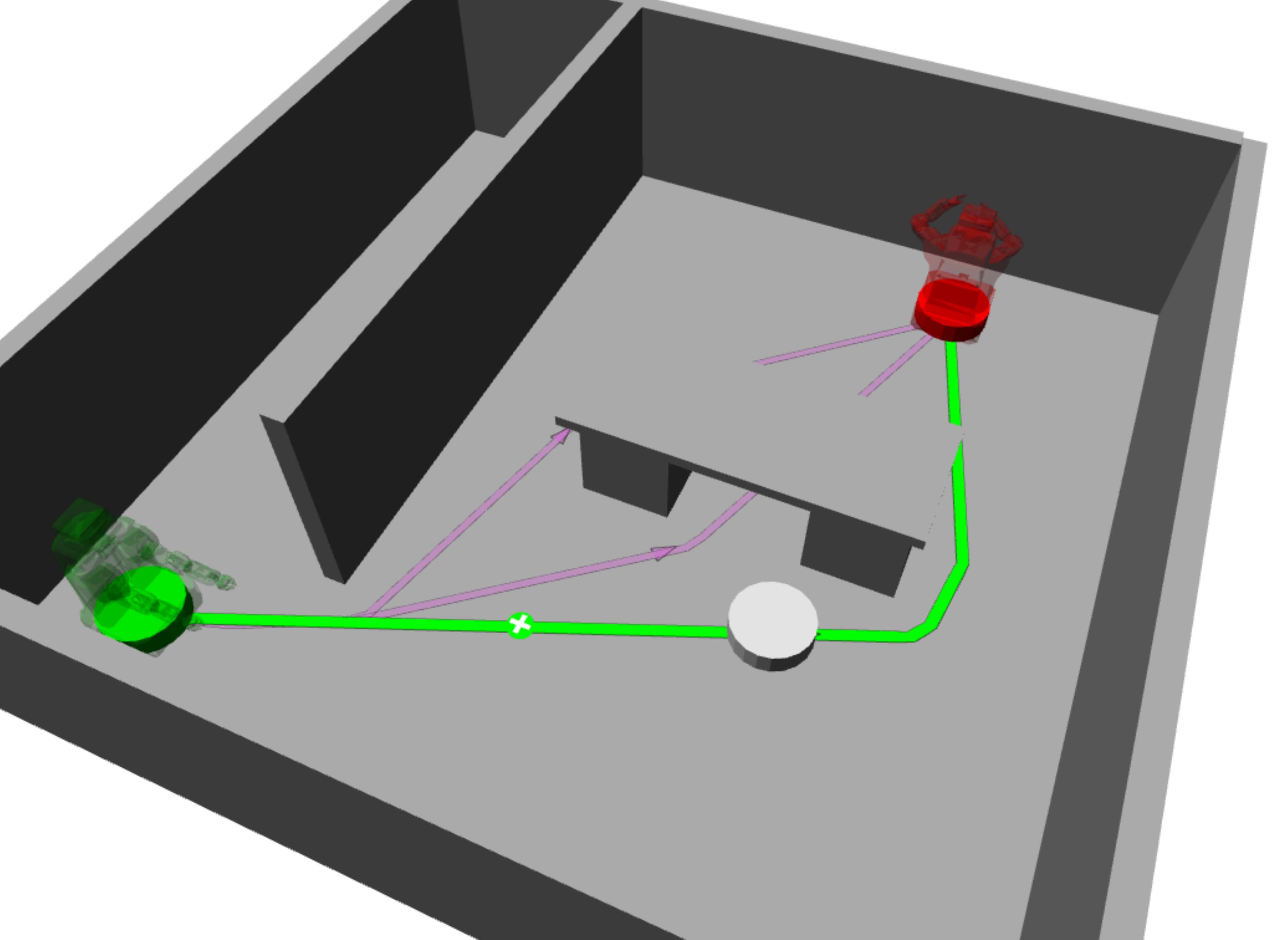}
    \includegraphics[width=0.32\textwidth]{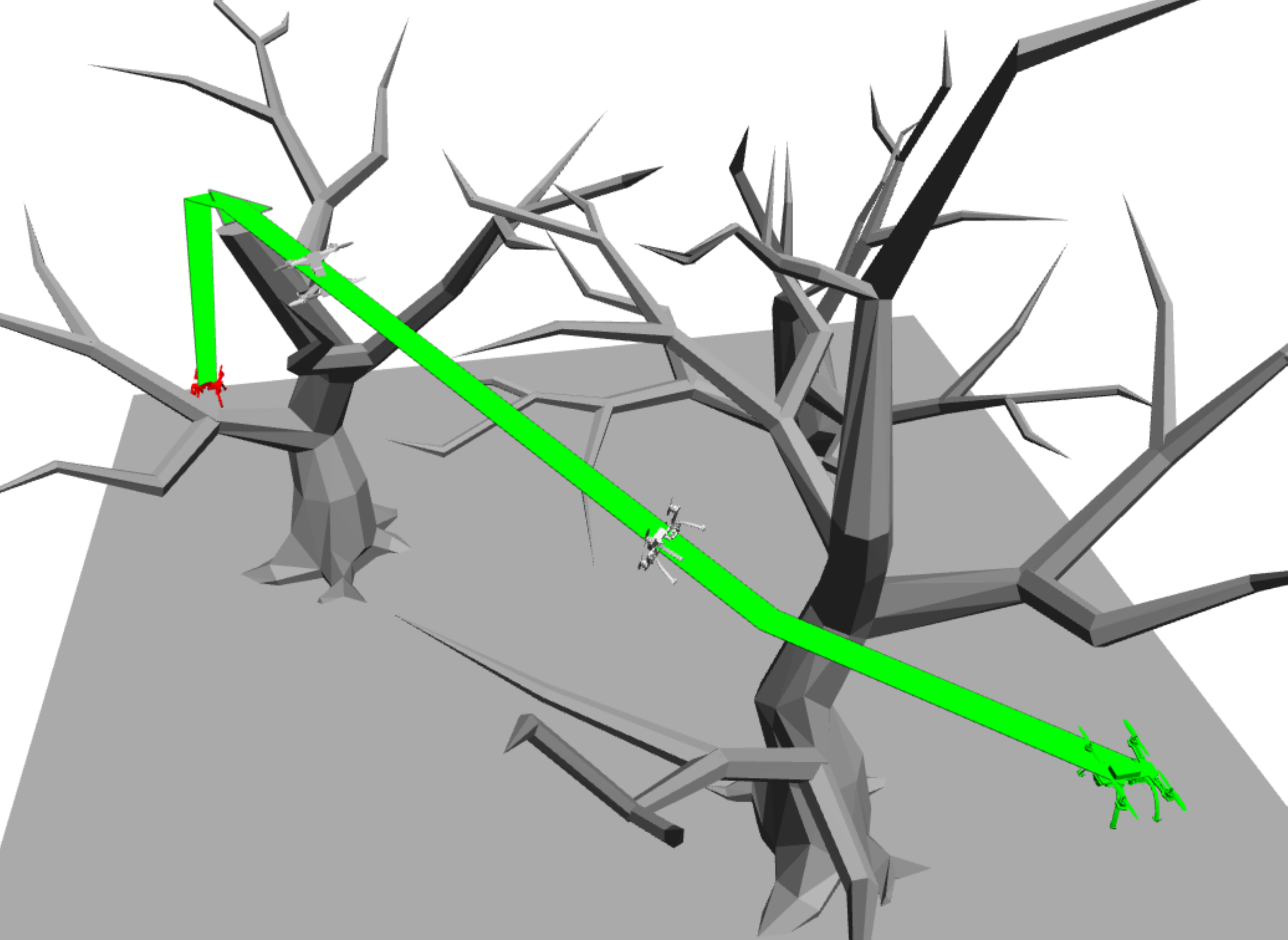}
    \includegraphics[width=0.32\textwidth]{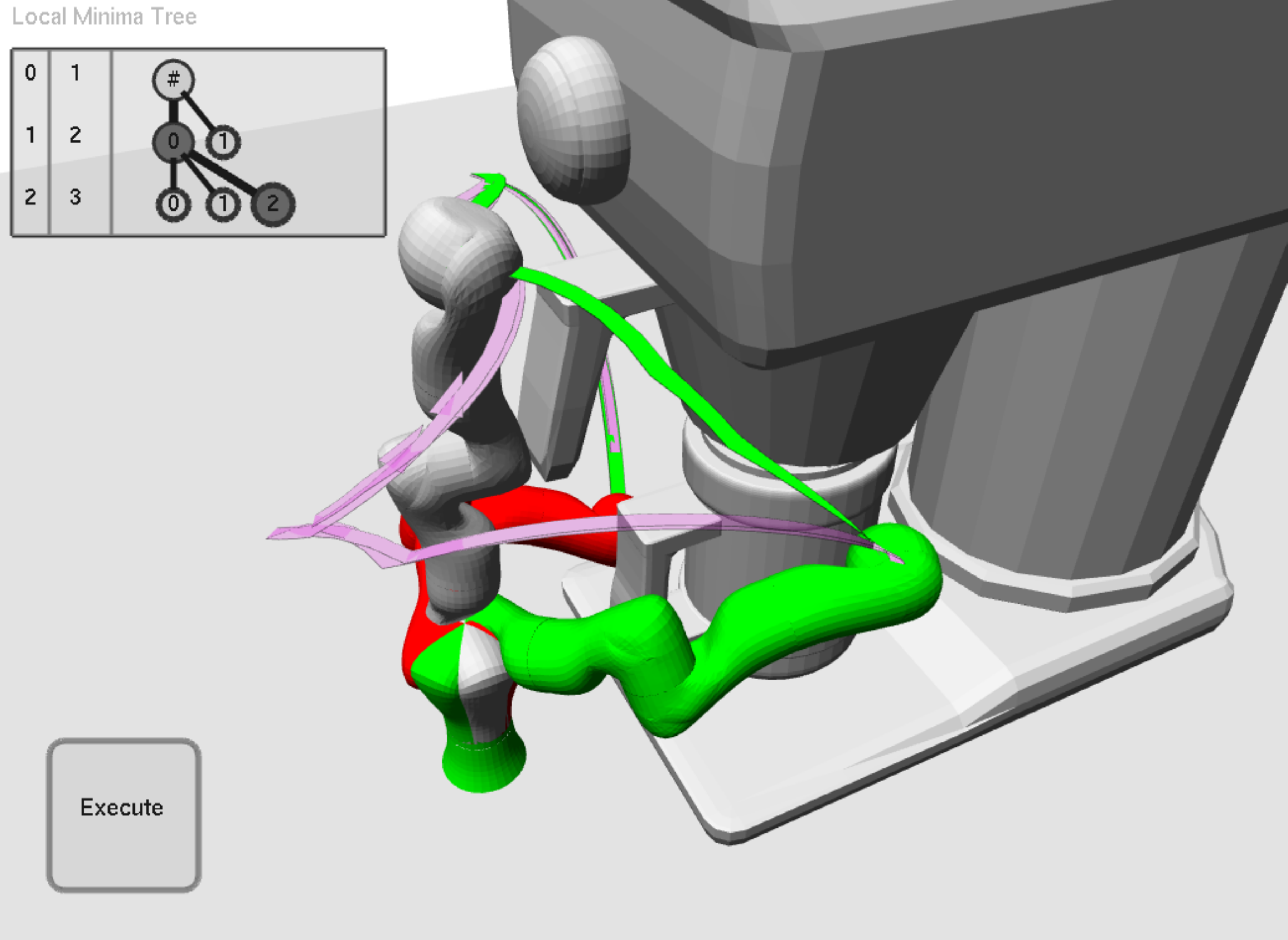}
    \includegraphics[width=0.32\textwidth]{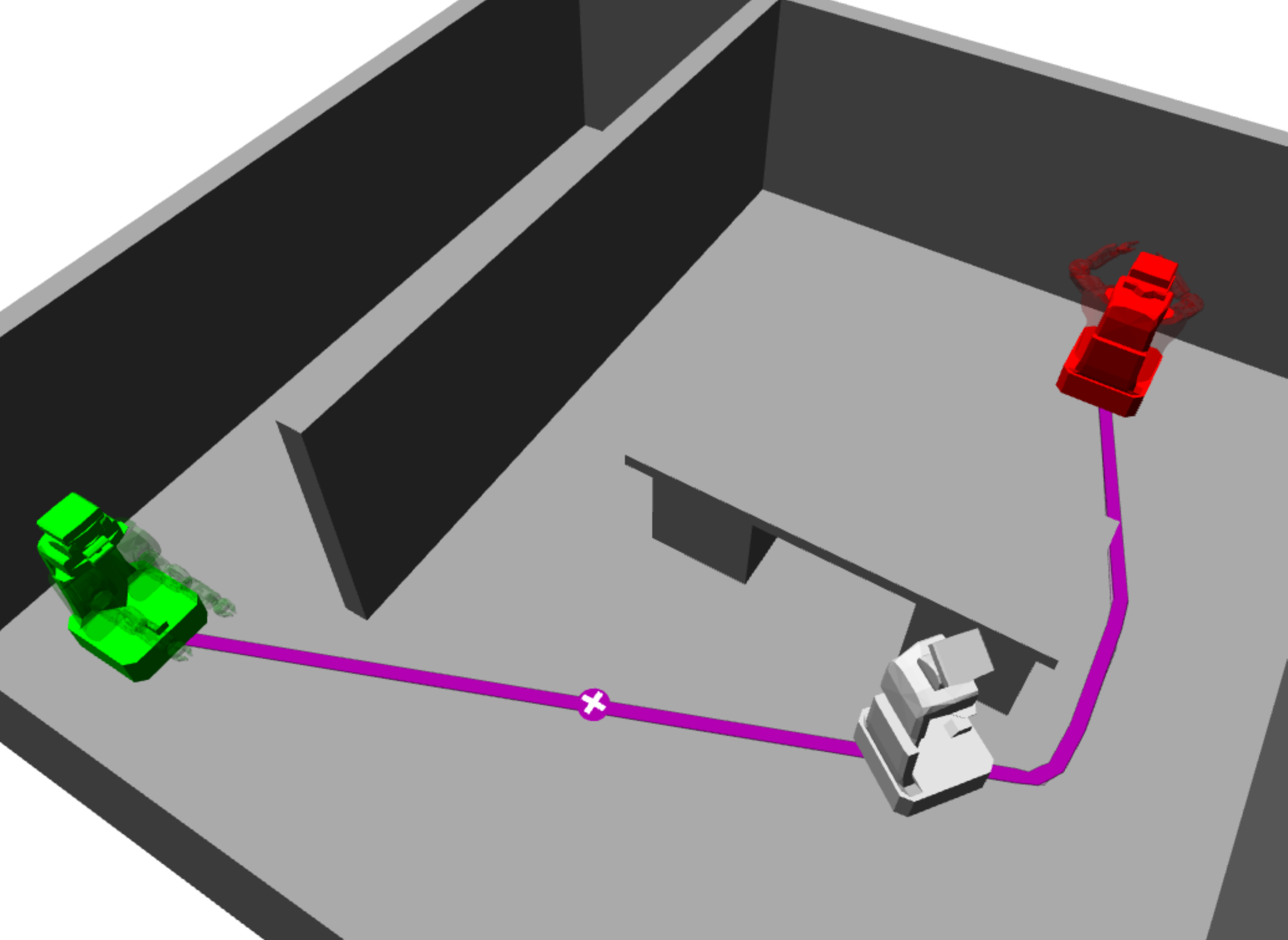}
    \caption{\textbf{Left}: 6-dof Drone \textbf{Middle}: 7-dof KUKA LWR \textbf{Right}: 34-dof PR2}
    \label{fig:experiment_geometric}
\end{figure*}
\begin{figure}[ht!]
    \centering
    \includegraphics[width=0.48\linewidth]{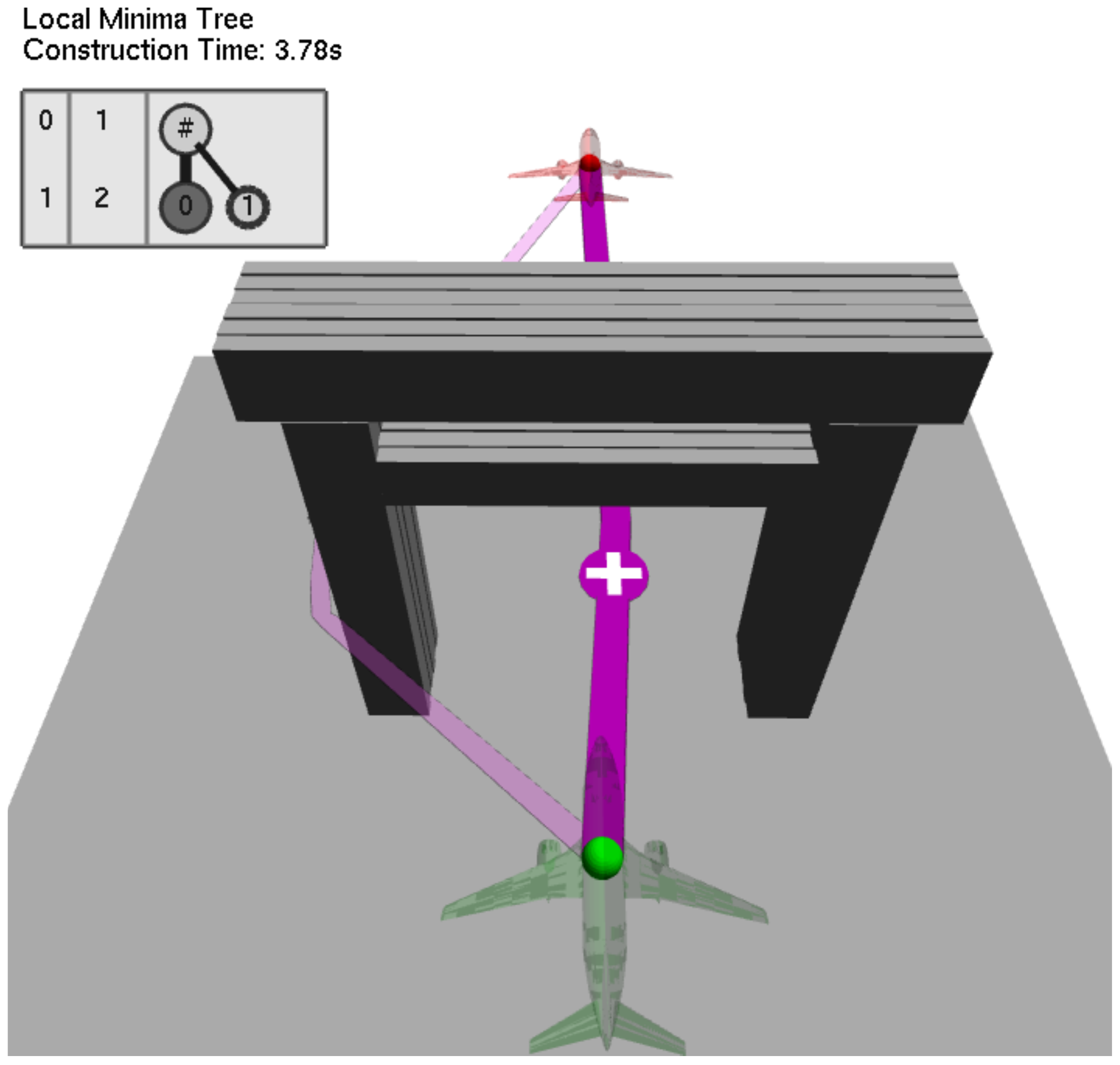}
    \includegraphics[width=0.48\linewidth]{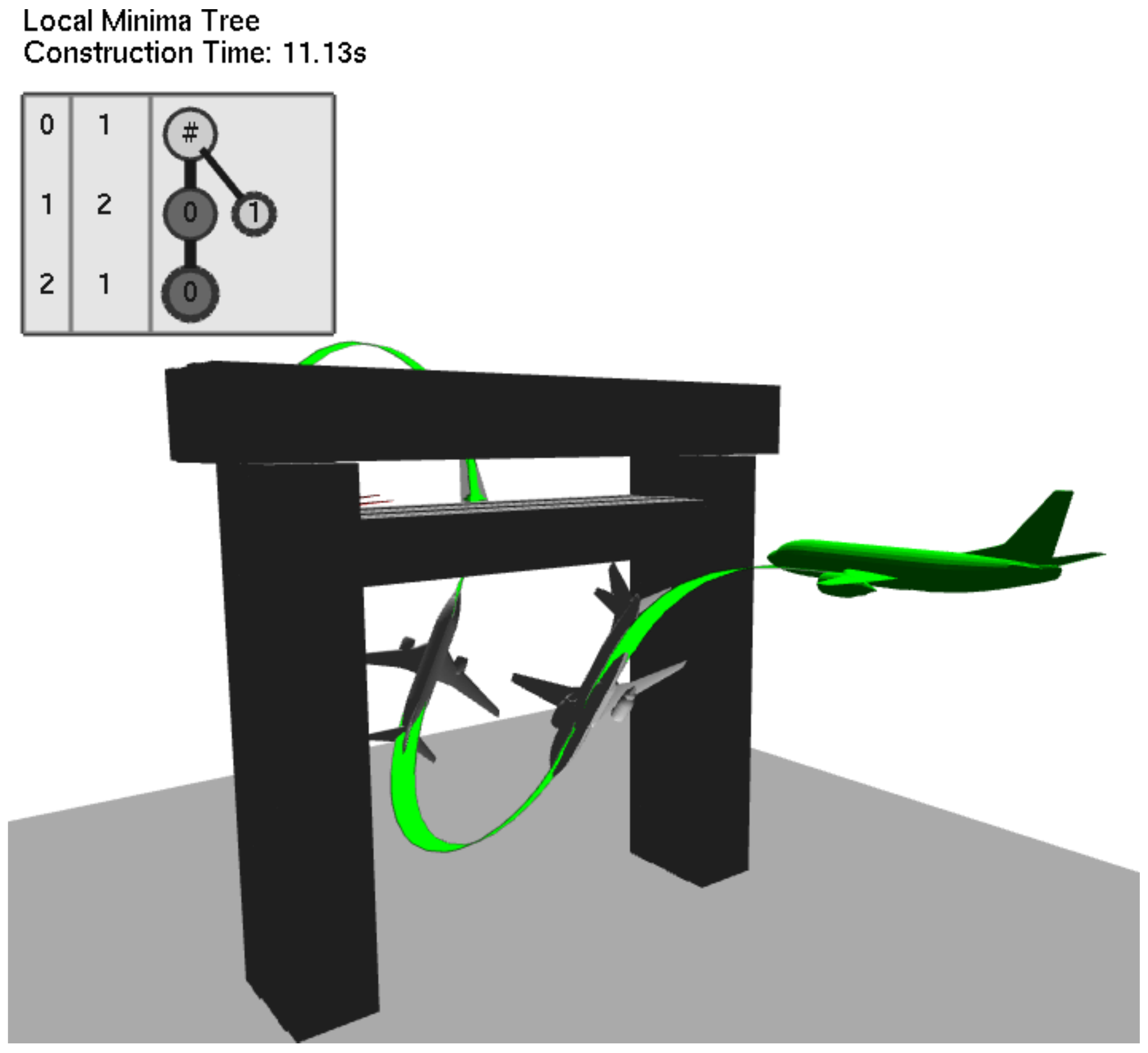}
    \caption{Dubin's airplane with
    constant forward velocity of $0.5$\siVelocity and bounds on first derivative of yaw and pitch of $\pm 0.1$\siVelocity. The dynamics are modelled as a driftless airplane \cite{orthey_2018b}.}
    \label{fig:experiment_airplane}
\end{figure}

\section{Demonstrations}

We demonstrate the motion planning explorer on four realistic and two pathological scenarios. We use a minimal-length cost function and a path optimizer implemented in OMPL\footnote{See \href{https://ompl.kavrakilab.org/classompl_1_1geometric_1_1PathSimplifier.html}{ompl::geometric::PathSimplifier}.}. For each configuration space, we pre-specify an admissible fiber bundle, based on runtime and meaningfulness of local-minima classes. In each scenario, we have used the parameters $N = 7$ (the maximum amount of visualized paths), the sparsity parameter $\deltaS = 0.1$ (the fraction of space visible from a vertex). We further set $\epsilon$ to $0.1$ times the measure of the space, and we have adjusted $\tmax$ to be between $1$s to $10$s. 
We perform each visualization on a $4 \times 2.50$GHz processor laptop using $8$ GB Ram and operating system Ubuntu $16.04$. 

We do not compare to existing methods, because we are not aware of any other algorithm which can (1) visualize local optima for any motion planning problem and (2) let a human user interact with it. 

For the four scenarios we have summarized the runtimes in Table \ref{tab:times}. The runtimes show the time to compute the local minima space $Q_0$ (Column 1), and the time to compute the remaining local minima spaces $Q_{>0}$ (Column 2) together with their sum (Column 3). Note that those times do not include the interaction by the user, and might differ depending on which minima have been selected. 

The first scenario is a drone in a forest. The configuration space is simplified using a fiber bundle $SE(3) \rightarrow \R^3$, corresponding to a sphere nested inside the drone. The outcome is shown in Fig. \ref{fig:experiment_geometric} (Left). The upper Figure shows seven local minima on the quotient space (magenta). Note that the quotient space is topologically trivial, but computing homotopical deformations would be computationally inefficient \cite{jaillet_2008}. 

The user selects the green path in phase one. We then compute local minima on the configuration space which project onto this path. In this case we find one single local-minima on $SE(3)$, which we then execute. 

Second, we use a robotic arm (Fig. \ref{fig:experiment_geometric} Middle) in an environment with a large coffee machine which has a visible geometric protrusion. The configuration is simplified using the fiber bundle $\R^7 \rightarrow \R^3$, obtained by removing the first three links of the robotic arm. The explorer finds two local minima on the quotient space which belong to a motion below the protrusion and above the protrusion, respectively. Finally, the user selects the path going above the protrusion, and the explorer finds three local minima which belong to different rotations of the manipulator around its axes.

Third, we use the PR2 robot in a navigation scenario. We use a fiber bundle $SE(2)\times \R^{31}  \rightarrow SE(2) \times \R^7 \rightarrow \R^2$, which corresponds to the removal of arms, and upper torso, respectively. On the lowest-dimensional quotient space, we find three local minima (Fig. \ref{fig:experiment_geometric} Right), which correspond to going left or right around the table, and one going underneath the table. Note that the path underneath the table is spurious (see Sec. \ref{sec:algorithm}). The computation of the first three local minima takes $4.61$s, while the remaining local-minima take together $292$s. This high runtime results from the high-dimensionality of the original configuration space combined with a possible narrow passage occurring when the robot has to traverse the corner of the table.

In the last scenario, we visualize the flight paths of dubin's airplane \cite{lavalle_2006} through an archway. Dubin's airplane is a rigid body in 3D with velocity constraints such that it flies at a constant forward velocity of $+0.5$\siVelocity and has bounds on the first derivative of yaw and pitch of $\pm0.1$\siVelocity. The fiber bundle is $SE(3) \times \R^6 \rightarrow \R^3$, which corresponds to the removal of dynamical constraints and orientation. We see that the algorithm finds two distinct local minima on the quotient space, which corresponds to going through or around the archway. We then select the local minimum going through the archway and the algorithm finds a dynamically feasible path (Fig. \ref{fig:experiment_airplane} Right). 

This demonstrates that our method can be extended to dynamical systems, even when the shortest path of the geometrical system is neither dynamically feasible nor near to a dynamically feasible path.
However, as detailed in Sec. \ref{sec:algorithm}, we currently do not have an adequate optimization function to compute a dynamically optimal path. The resulting dynamical path is therefore non-optimal.

\begin{table}[t]
    \centering
    \begin{tabular}{|c|c|c|c|}
        \hline
        Scenario  & Time $Q_0$ (s) & Time $Q_{>0}$ (s) & Total Time (s)\\
        \hline
        Planar Manipulator & $0.51$&$0.82$& $1.33$\\
        Planar Car & $1.57$ & $3.16$ & $4.73$\\
        Drone in Forest & $9.57$ & $0.89$ & $10.46$\\
        Robotic Arm & $2.03$ & $10.57$ & $12.6$\\
        PR2 &  $4.61$ & $292.29$ & $296.9$\\
        Dubin's Airplane & $2.15$ & $12.34$ & $14.49$\\
        \hline

    \end{tabular}
    \caption{Time (s) to generate the local-minima tree in the demonstration cases shown in Video.}
    \label{tab:times}
\end{table}

\subsection{Pathological Scenarios and Limitations}

While the motion planning explorer works well on realistic scenarios, it might not work well on pathological cases. To test this, we demonstrate the performance on two scenarios which have been crafted to break the algorithm. 

\begin{figure}[h!]
    \centering
    \includegraphics[width=0.44\linewidth]{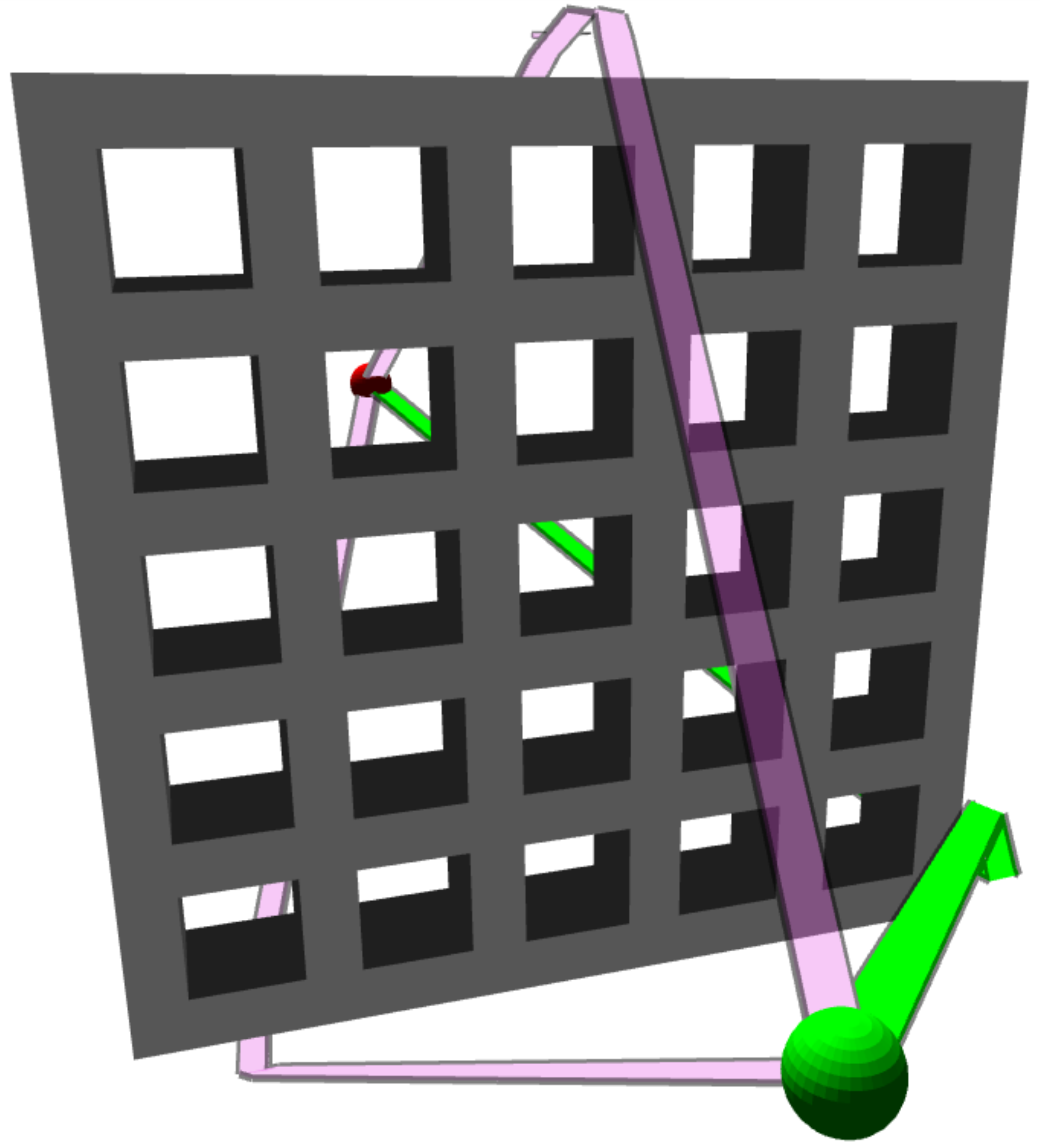}
    \includegraphics[width=0.54\linewidth]{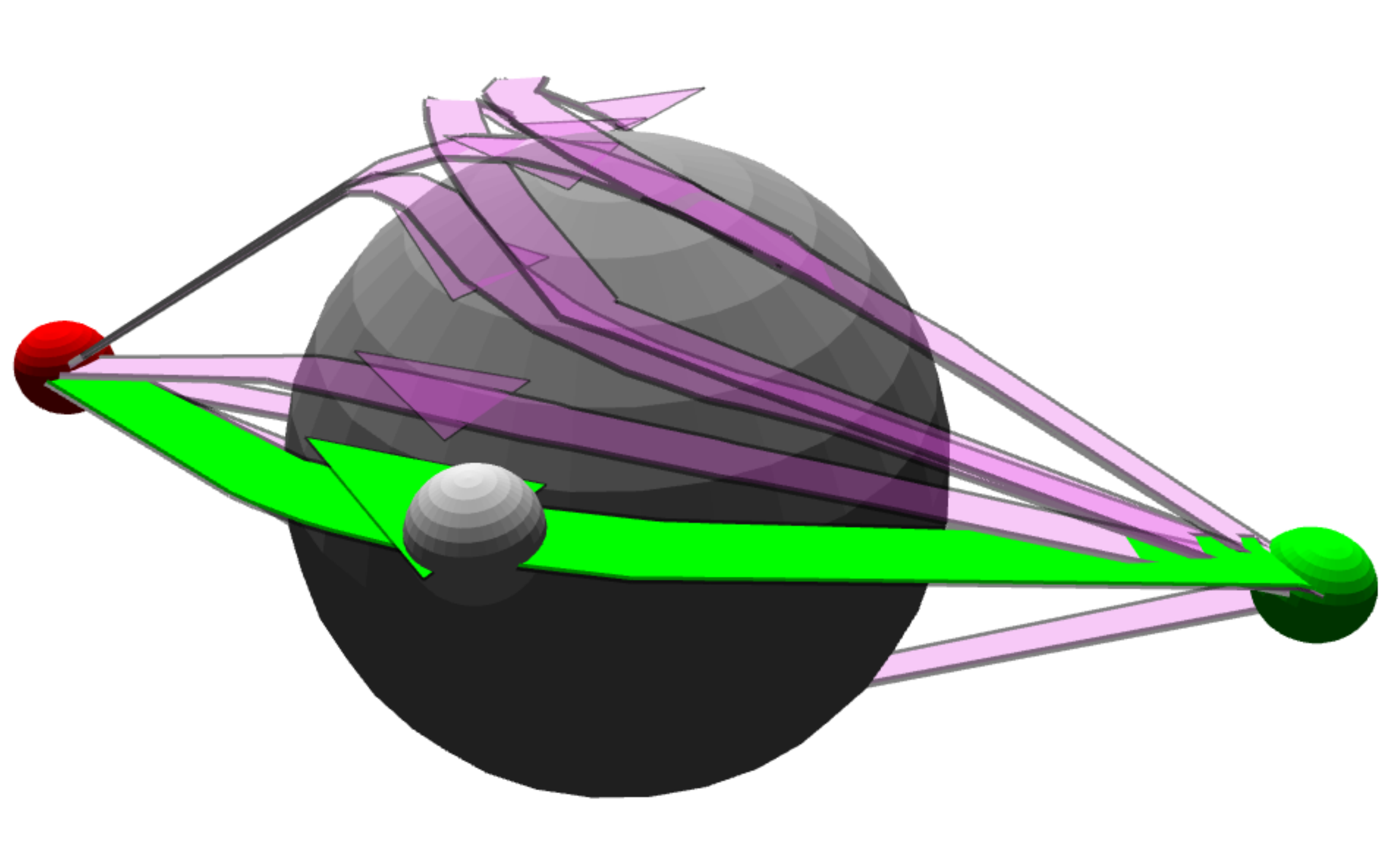}
    
    \caption{Limitations: (A) low-cost small-measure local-minima are often ignored in favor of high-cost but large-measure local-minima. 
    (B) Only a countable number of paths is found from an uncountable number of local minima.}
    \label{fig:limitations}
\end{figure}

In the first scenario (Fig. \ref{fig:limitations} Left), we need to move a ball
with configuration space $\R^3$ from an initial (green) to a goal (red)
configuration. Between the configurations we place a lattice with openings
slightly larger than the radius of the ball. All local minima through the lattice
have a neighborhood in pathspace with a vanishingly small measure. Our
algorithm, however, rarely detects those minima, because the probability of finding samples inside an opening is smaller than finding samples above or below the lattice. Therefore, the algorithm usually finds minima with a higher cost going around the lattice. 

In the second scenario (Fig. \ref{fig:limitations} Right), we place a spherical
obstacle between the initial and goal configuration of the ball (As described by Karaman and Frazzoli
\cite{karaman_2011}). The number of local minima is uncountable infinite. Our
algorithm, however, can only find a finite number of local minima and is unable to describe the
complete uncountable set. 

\section{Conclusion}

We introduced the motion planning explorer, an algorithm taking a planning
problem as input and computing a local-minima tree. Local minima are
defined as paths which are invariant under minimization of a cost functional. The local-minima are grouped into a tree, where two
paths are grouped together if they are projection-equivalent under a
lower-dimensional fiber bundle projection.  We showed that the resulting
local-minima tree faithfully captures the structure of holonomic and certain non-holonomic problems.

The implementation of the local-minima tree has, however, three limitations. First, we restrict computation to $N$ simple paths, which makes the tree non-exhaustive. We could alleviate this by letting the user add additional local-minima and by enumerating non-simple paths. Second, the runtime is sometimes prohibitive for real-time application. We believe this could be addressed by specifically tailored hardware
\cite{murray_2016}, code optimization and a sampling-bias towards narrow passages. Third, the construction of the tree depends on pre-specified lower-dimensional projections. We could remove this dependency by enumerating all projections \cite{orthey_2019} and use a specific projection only if it will group at least two local minima together. 

Most importantly, however, the computation time spent constructing the local-minima tree
is negligible compared to having a tool which allows us to visualize, debug and
interact with a planning problem.

\section{Acknowledgements}

We thank the anonymous reviewers for clarifying the connection to Morse theory. This paper was supported by the Alexander von Humboldt foundation. We thank Marc Moll and Zachary Kingston
for independent code reviews and the website
\href{www.turbosquid.com}{TurboSquid} for providing 3D models.

\bibliographystyle{IEEEtranS}
{\small
\bibliography{IEEEabrv,bib/quotient-space,bib/planner,bib/general,bib/pathdeformation}
}
\end{document}